%% file: main.tex
\pdfoutput=1

\documentclass[11pt]{article}

\usepackage[final]{acl}

\usepackage{times}
\usepackage{latexsym}

\usepackage[T1]{fontenc}

\usepackage[utf8]{inputenc}

\usepackage{microtype}

\usepackage{inconsolata}

\usepackage{graphicx}

\usepackage{amsmath}
\usepackage{listings}
\usepackage{algorithm}
\usepackage{algorithmic}
\usepackage{caption}
\usepackage{subcaption}
\usepackage{pythonhighlight}
\usepackage{booktabs}
\usepackage{enumitem}
\usepackage{float}
\usepackage{stfloats}
\usepackage{wrapfig}

\title{\methodname{}: Reasoning with Intermediate Revision and Search}

\author{Yizhou Chi \and Kevin Yang \and Dan Klein \\
UC Berkeley\\
\texttt{\{yizhouchi,yangk, klein\}@berkeley.edu} \\
}

\newcommand{\methodname}{\textsc{ThoughtSculpt}}
\begin{document}
\maketitle
\input{latex/sections/00-abstract}
\input{latex/sections/10-intro}

\input{latex/sections/20-related}

\input{latex/sections/30-methods}

\input{latex/sections/40-experiments}

\input{latex/sections/50-discussion}

\input{latex/sections/60-Reproducibility}

\bibliography{latex/bib/story, latex/bib/refine, latex/bib/feedback, latex/bib/experiments, latex/bib/planning}
\newpage
\onecolumn
\input{latex/sections/99-Appendix}
\end{document}

%% file: latex/sections/00-abstract.tex
\begin{abstract}
We present \methodname{}, a general reasoning and search method for tasks with outputs that can be decomposed into components. \methodname{} explores a search tree of potential solutions using Monte Carlo Tree Search (MCTS), building solutions one action at a time and evaluating according to any domain-specific heuristic, which in practice is often simply an LLM evaluator. Critically, our action space includes revision actions: \methodname{} may choose to revise part of its previous output rather than continuing to build the rest of its output. Empirically, \methodname{} outperforms state-of-the-art reasoning methods across three challenging tasks: Story Outline Improvement (up to +30\% interestingness), Mini-Crosswords Solving (up to +16\% word success rate), and Constrained Generation (up to +10\% concept coverage). 
\end{abstract}

%% file: latex/sections/10-intro.tex
\section{Introduction}

While large language models (LLMs) such as GPT \citep{brown2020language, openai2024gpt4}, LLaMA \citep{touvron2023llama, touvron2023llama2}, and Claude \citep{Anthropic_2024} are increasingly capable at performing a variety of reasoning tasks, recent studies have revealed that the utilization of distinct prompting strategies and instructional guidance can have a notable influence on the performance of LLMs when tackling identical tasks.

Chain-of-Thought (CoT) is a prompting strategy detailed in \citet{wei2023chainofthought} that directs LLMs to produce the final task output through intermediate steps of reasoning, referred to as "intermediate thoughts." Notably, CoT has demonstrated a substantial enhancement in the problem-solving proficiency of LLMs without necessitating any model updates. Self-consistency with CoT (CoT-SC) \citep{wang2023selfconsistency} proposes to improve output consistency by generating multiple CoTs and selecting the best outcome. Recently, extending CoT and CoT-SC, Tree-of-Thoughts \citep{yao_tree_2023} and Graph-of-Thoughts \citep{besta2024graph} propose to shape the reasoning process of LLMs as a tree or an arbitrary graph structure. These approaches enable LLMs to explore different paths of thought and find better outputs by utilizing backtracking and graph-search algorithms. However, these approaches' reasoning capabilities are often limited by the set of candidates they generate at earlier steps. They cannot revise and edit their original answers continuously in later steps. As a result, these methods may not be as effective in addressing problems that require frequent revision and modifications. 

We propose \methodname{}, a tree-based framework that emulates human reasoning by enabling LLMs to create interconnected thought networks. A key feature is its self-revision mechanism, which iteratively improves outputs while generating new thought nodes. To address the vast search space in text generation, we use Monte Carlo Tree Search (MCTS), which efficiently navigates the search space and provides high-quality solutions, though not necessarily globally optimal. Our method includes three core modules: the thought evaluator, which gives textual and numerical feedback; the thought generator, which produces solutions based on initial instructions and feedback; and the decision simulator, which simulates lines of thought within the MCTS process to assess the potential value of different paths.

\begin{figure*}[htbp]
\includegraphics[width=\textwidth]{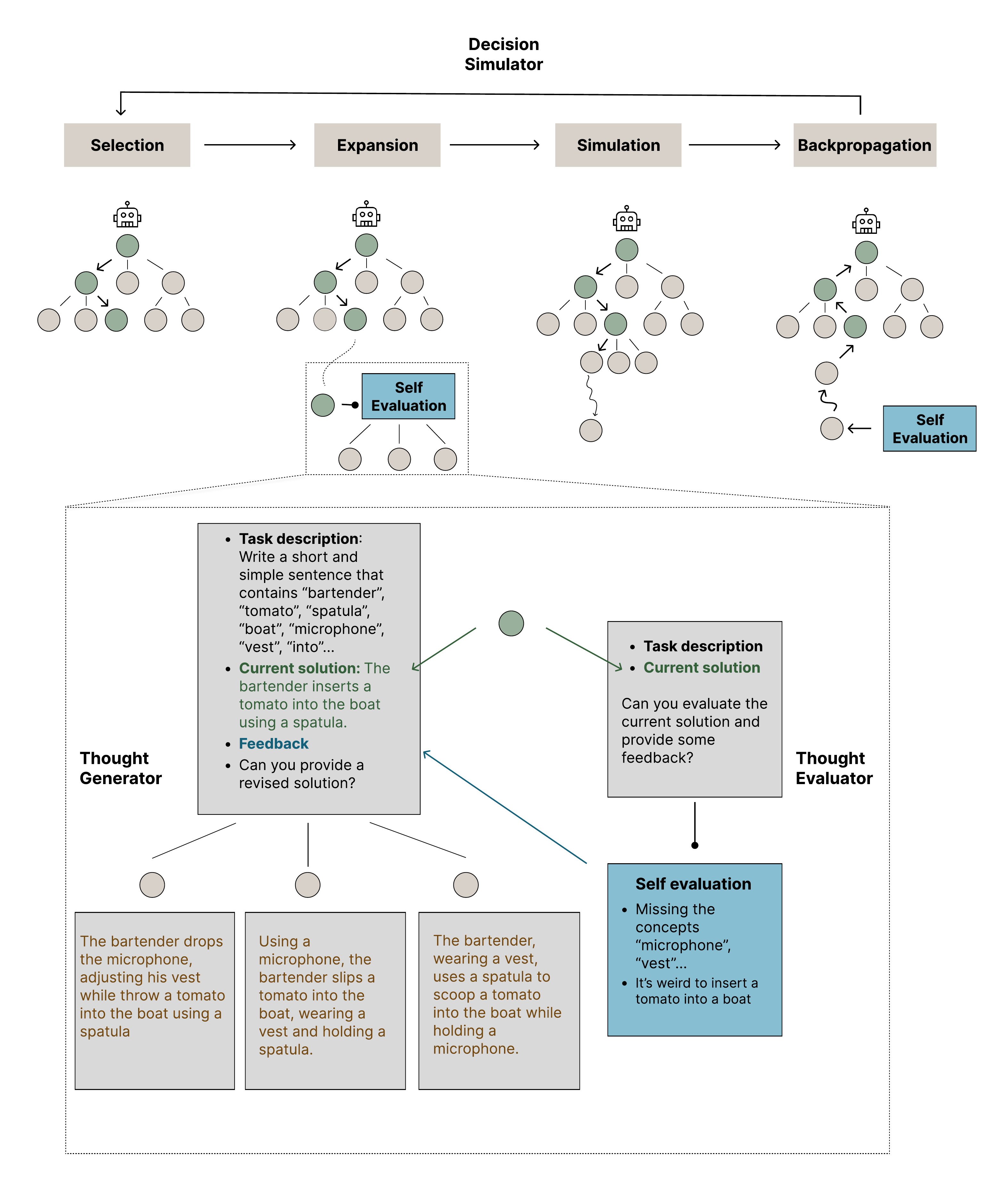}
\caption{\small Illustration of \methodname{} using Monte Carlo Tree Search on the Constrained Generation task. Each circle in the diagram represents a thought node generated by LLMs.  \textit{Selection:} choose a thought node $x$ based on a selection algorithm. \textit{Expansion:} A new set of child nodes $X$ is generated using the initial instruction, the current node, and self-evaluated textual feedback. The zoom-in of the expansion phase demonstrates the use of the \textbf{Thought Evaluator} and the \textbf{Thought Generator}, which entails assessing and refining the current solution for the task \ref{task3}. \textit{Simulation:} a single node $x'$ is randomly chosen from the set $X$. This selected node $x'$ generates further nodes in sequence for several steps, corresponding to our \textbf{Decision Simulator}. \textit{Backpropagation:} The numerical feedback evaluated at the last node is propagated back to the root node.}
\label{main-diagram}
\end{figure*}


We evaluate \methodname{} on three challenging tasks for state-of-the-art language models: Story Outline Improvement, Mini-Crosswords Solving, and Constrained Generation. These tasks require advanced reasoning skills, varying degrees of exploration, and the ability for self-revision to achieve optimal results. Compared to state-of-the-art reasoning strategies as baselines, \methodname{} exhibits an up to 30\% interestingness increase in Story Outline Improvement; up to 16\% word success rate increase in Mini-Crossword Solving; and up to 10\% concept coverage improvement in Constrained Generation. These findings underscore the efficacy of \methodname{} across diverse tasks.


%% file: latex/sections/20-related.tex
\section{Related Works}

\paragraph{Feedback Guided Generation.}
Human feedback has been shown to be effective in improving LLMs' generation \cite{tandon2022learning, elgohary2021nledit, bai2022training}. However, human feedback is often costly and unable to be incorporated into an automated generation process. As a result, some works adopt a heuristic function to serve as an alternative to human feedback \citep{liu2022rainier, lu2022quark, le2022coderl, welleck2022generating}.

\citet{madaan_self-refine_2023, shinn_reflexion_2023, paul_refiner_2024} introduce a mechanism for LLMs to produce self-reflective feedback to improve their outputs. Along with the model-generated feedback, \citet{chen2023teaching} uses execution results to help improve code generation. Likewise, \citet{kim2023language} introduces a critic step to improve the model's performance in computer tasks. These approaches follow left-to-right linear processes, potentially overlooking alternative directions. In our work, each thought node having multiple children nodes allows for broader exploration, enhancing decision-making comprehensiveness.

\paragraph{Graph Reasoning.}
To facilitate broader exploration in problem-solving, \citet{yao_tree_2023} and \citet{xie2023selfevaluation} use a tree-search procedure where each node represents a partial solution, requiring a complete solution to combine multiple nodes. This method restricts modifications to intermediate nodes, making the final output reliant on initial candidates. \citet{besta2024graph} proposed a graph-based paradigm that models LLM reasoning as an arbitrary graph, allowing combinations of connecting nodes. Our approach differs by permitting review and modification of intermediate nodes, even allowing them to be revised or expanded if initially complete. This flexibility improves expressivity and enables language models to correct initial mistakes. Several concurrent works \cite{hui2024rotenhancinglargelanguage, tian2024selfimprovementllmsimaginationsearching, chen2024alphamathzeroprocesssupervision} have recently explored integrating Monte Carlo Tree Search (MCTS) with Large Language Models (LLMs). However, these approaches primarily focus on mathematical reasoning tasks or rely on external feedback, fine-tuned policies, or reward models. In contrast, our method operates entirely at inference time, requiring no additional model training or external feedback.

\paragraph{LM Planning.}
Long-form generation and complex problem-solving often require high-level planning or outlining. Natural language outliners and structured schemas play integral roles in generating long-form content \citep{tian-peng-2022-zero, mirowski2022cowriting,yang_re3_2022, yang_doc_2023}. There are also works that utilize LLMs to tackle complex tasks such as video games, fact-checking, housekeeping, and code optimization with planning using natural languages \citep{yao2023react, huang2022language, wang2023describe, huang2022inner}. Our work could also be seen as a generic task planner using LLMs that leverages Monte Carlo Tree Search to facilitate various tasks in diverse domains.

%% file: latex/sections/30-methods.tex
\section{Method}
We treat each formal output of LMs as a thought node $x \in \{x^{0}, x^{1}, ... x^{i}\}$, where $x^{0}$ is the root node and the initial output provided by LMs given the task instruction $I$. For instance, a thought node can be a few lines of items (Story Outline Improvement), a couple of words (Mini-Crosswords), or a sentence (Constrained Generation). To process the thought node and look for a better output, our method consists of three modules: thought evaluator, thought generator, and decision simulator. 

\begin{figure*}[h]
\includegraphics[width=0.9\textwidth]{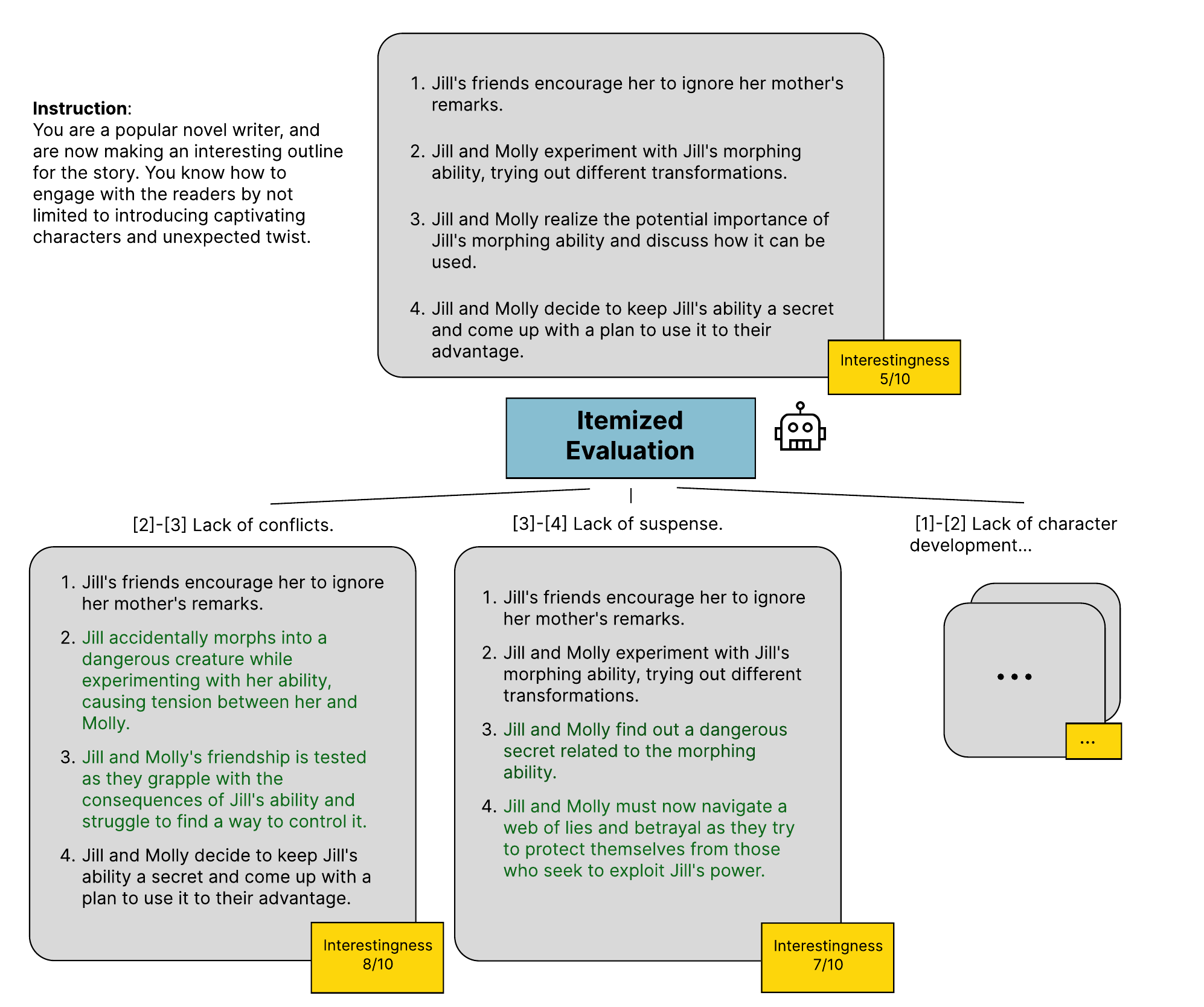}
\caption{\small Illustration of our Story Outline Improvement task. A step involves employing the thought evaluator to conduct itemized evaluations of the story outline and utilizing the thought generator to generate a candidate set of improved story outlines for task \ref{task1}.}
\label{fig:outline}
\end{figure*}

\subsection{Thought Evaluator}
The thought evaluator evaluates the status of each thought node and provides feedback for potential improvement. It not only works as a heuristic for the search algorithm but also gives potential directions and guidance to generate new candidates. 

Feedback $f(x^{i})$ for a node $x^{i}$ consists of numerical feedback $f_{numeric}(x^{i})$ and natural language feedback $f_{NL}(x^{i})$. The numerical feedback will be used as the evaluation score $v(x^{i})$ for the current node, and the natural language feedback will be used as context to generate child nodes.

\begin{align}
f(x^{i}) =\enspace <f_{NL}(x^{i})&, f_{numeric}(x^{i})> \\
f_{numeric}(x^{i}) &= v(x^{i})
\end{align}

We present two types of natural language feedback, each beneficial for various task scenarios. Furthermore, these strategies are flexible, allowing for independent or combined utilization.
\begin{itemize}[topsep=0pt,itemsep=-1ex,partopsep=1ex,parsep=1ex]
    \item \textit{Holistic Evaluation:} Evaluate the entire thought node as a unified whole to provide comprehensive feedback. This approach captures the core message and coherence of the node. The right side of the zoomed-in expansion phase in Figure \ref{main-diagram} illustrates how the thought evaluator generates holistic feedback based on the task description and the current solution of the node.

    \item \textit{Itemized Evaluation:} Evaluate each sub-unit of the thought node individually, providing targeted feedback for each component. This method results in a list of feedback specific to each sub-unit, making it ideal when the thought node can be divided into distinct elements for localized evaluation. For instance, in the story outline task shown in Figure \ref{fig:outline}, breaking the outline into separate items allows for focused assessment and refinement.
\end{itemize}


\subsection{Thought Generator}

Once we have evaluation feedback of the current node, we can form subsequent thought nodes that aim to improve the current output. Based on the task description $I$, the current solution $x_{parent}$, and the natural language feedback $f_{NL}$ provided by the self-evaluator, each thought node generates $k$ candidate thought nodes using a pre-trained LM with a parameter $\theta$. 

A child node $x_{child}$ will be generated as follows:
\begin{align}
x_{child} \sim p_{\theta} (x | I, x_{parent}, f_{NL}(x_{parent}))  
\end{align}

The left part of the zoomed-in expansion phase depicted in Figure \ref{main-diagram} illustrates how \methodname{} leverages the task description, current solution, and evaluation feedback to produce a set of candidate nodes.

\subsection{Decision Simulator}
\methodname{} is equipped with a decision simulator that enables it to simulate decisions at deeper layers and then backpropagate to update the score of the current decision. In other words, we are doing a rollout to get a better estimate of the reward for the node we are at. The behavior of the decision simulator is analogous to the processes in Monte Carlo Tree Search (MCTS; see Algorithm \ref{alg:mcts}). It is possible to replace the decision simulator with other search algorithms such as DFS, BFS, or A* search (and we in fact run DFS as well in our experiments in Section \ref{sec:experiments}), but MCTS provides a computational advantage by efficiently navigating complex search spaces, balancing exploration and exploitation to reach optimal solutions with fewer evaluations. Its incremental and iterative nature also scales well to large problem instances.

MCTS explores potential moves and stores the outcomes in a search tree. With each search iteration, the tree expands, accumulating more information. As shown in Figure \ref{main-diagram}, MCTS can be divided into four phases: selection, expansion, simulation, and backpropagation. 

In the selection phase, a leaf node will be selected based on Upper Confidence Bound 1 (UCB1) Eqn \ref{eq:UCB1} which prioritizes nodes that have not been explored extensively but show promise. Therefore, the UCB1 value of node $x$ takes into account not only the heuristic score $v(x)$ but also the total number of visits to the node itself, $n(x)$, as well as its parent node, $n(x_{parent})$.

\begin{align}
\label{eq:UCB1}
UCB1(x) = v(x) + c \sqrt{\frac{\ln n(x_{parent})}{n(x)}}
\end{align}

In the expansion phase, the thought generator will expand the selected leaf node by generating a set of children nodes based on the feedback provided by the thought evaluator. 

In the simulation phase, a child node is picked from the newly generated set using a uniform distribution. In the subsequent iterations, however, we generate only a single node iteratively until the maximum simulation depth $d_{simulation}$ is reached. 

Finally, in the backpropagation phase, we update the reward of the last node generated in the simulation back to the root node and iterate this process for $d_{rollout}$ steps. The node with the highest average reward will be chosen as the final output.

%% file: latex/sections/40-experiments.tex
\section{Experiments}
\label{sec:experiments}

We evaluate our method on three distinct tasks: Story Outline Improvement, Mini-Crossword Solving, and Constrained Generation. 

We evaluate the tasks with Chain-of-Thought (CoT) \citep{wei2023chainofthought}, Self-Refine \citep{madaan_self-refine_2023}, and Tree-of-Thoughts (ToT) with DFS \citep{yao_tree_2023} as baselines. We use GPT-3.5 (\texttt{gpt-3.5-turbo-0125}) and GPT-4 (\texttt{gpt-4-0125-preview}) \citep{openai2024gpt4} as strong base LMs for the reasoning algorithms across all tasks. Both base LMs use a temperature of 0.7. To further evaluate the efficacy of our proposed approach, we conduct an ablation study by investigating the performance of our method when employing Depth-First Search (DFS) (Algorithm \ref{alg:dfs_code}) as an alternative search algorithm to the MCTS algorithm. 
In addition, running \methodname{} with DFS facilitates closer comparison with ToT, which also uses DFS. While \methodname{} with MCTS typically performs better, we observe in our experiments below that \methodname{} with DFS still outperforms our other baselines, demonstrating \methodname{}'s ability to generalize to other search algorithms. 

\subsection{Story Outline Improvement}
\label{task1}

\begin{table}[!htbp]
\centering
        \begin{tabular}{@{}lcc@{}}
            \toprule
            & \multicolumn{2}{c}{\textbf{\textit{Base LLM}}}  \\ \midrule
            \textbf{Methods}     & \textbf{GPT3.5}    & \textbf{GPT4}     \\ \midrule
            Initial Outline & 12.0 & 12.0 \\ \midrule
            CoT                 & 50.1 & 28.8 \\
            Self-refine         & 65.5 & 27.9   \\
            ToT                 & 72.1 & 49.9    \\
            \methodname{} (DFS) & 79.3 & 53.7 \\ 
            \methodname{} (MCTS) & \textbf{89.9} & \textbf{65.0} \\ \bottomrule
            \end{tabular}
        \captionof{table}{Average outline interestingness. Initial Outline is the starting point before rewriting with any reasoning method. \methodname{}'s outputs are judged to be interesting at a higher percentage compared to baselines.}
        \label{table:outline}
\end{table}


One approach to generating long-form stories via LLMs is to adopt a high-level writing process that first designs an outline of the story and fills up the details based on the outline \citep{yang_re3_2022, yang_doc_2023}. An unengaging or uncompelling outline is unlikely to yield a captivating final draft, regardless of the subsequent detailing efforts. To address this challenge, we propose a task focused specifically on enhancing the interestingness of story outlines generated by LLMs. 

\paragraph{Task Setup} 
We sample 500 book descriptions from the WhatsThatBook dataset \citep{lin2023decomposing} and generate story outlines using DOC \citep{yang_doc_2023} with GPT-3.5. We allocate 400 descriptions for training, 50 for validation, and 50 for testing. For each description, we generate three types of outlines: one prompted to be interesting, one prompted to be boring, and one without specific instructions. Since there is no ground truth for the interestingness of the outline, we employ an outline content evaluator to assess the final interestingness of generated or revised outlines. Neither \methodname{} nor the baselines have access to this evaluator during outline generation. We fine-tune the pre-trained Flan-T5 model \citep{chung2022scaling} to serve as the content evaluator, training it to rate interesting outlines as 1 and boring ones as 0. This evaluator's output serves as the score metric for the task. For evaluation, LMs revise and improve the interestingness of default outlines in the test set. The dataset includes 400 interesting and 400 non-interesting outlines for fine-tuning, 50 interesting and 50 non-interesting outlines for validation, and 50 outlines for testing algorithms. 

Also, we conduct human evaluation via Prolific to assess the generated story outlines (GPT-3.5 as the base LLM), capturing subjective perceptions and cultural nuances that LLMs may miss. We recruited annotators to evaluate 100 pairs of story outlines, each pair consisting of one outline generated by \methodname{} with MCTS and another by ToT or Self-Refine, with each pair annotated by two annotators.
\begin{figure}
    \includegraphics[width=0.75\linewidth]{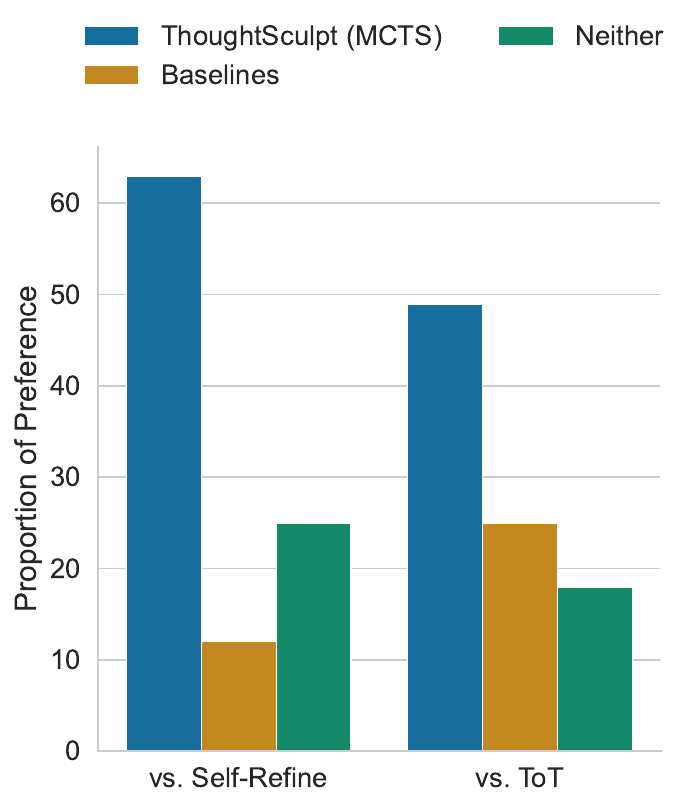}
        \centering
        \captionof{figure}{Proportion of outlines generated by each method that were preferred by humans in pairwise comparison. ("Neither" indicates that neither \methodname{} nor the baseline methods were preferred.)}
        \label{fig:humaneval}
\end{figure}

\paragraph{Method Setup}
Each method is allowed to search or iterate through a maximum depth of 3. The thought evaluator will perform an itemized evaluation on the current outline and provide an interesting score from 1 to 10 as the numerical feedback. Based on each itemized feedback, a child node will be proposed to modify the current outline in order to improve its interestingness. For \methodname{} and ToT, each node will generate a maximum of 3 candidate child outlines. In this and all the experiments below, \methodname{} with MCTS will have a maximum $d_{simulation}$ of 1. Figure \ref{fig:outline} illustrates how the story outline is improved.

\begin{figure}
    \includegraphics[width=0.9\linewidth]{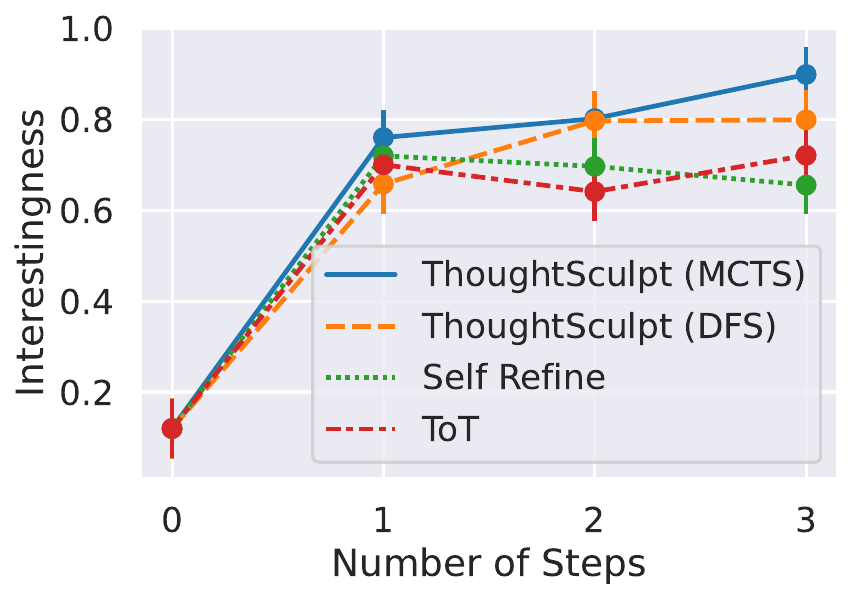}
        \centering
        \captionof{figure}{Average outline interestingness at each step. \methodname{}'s interestingness increases more with steps compared to baselines.}
        \label{fig:interestingness_cont}
\end{figure}
\begin{figure*}[!hbtp]
\includegraphics[width=\textwidth]{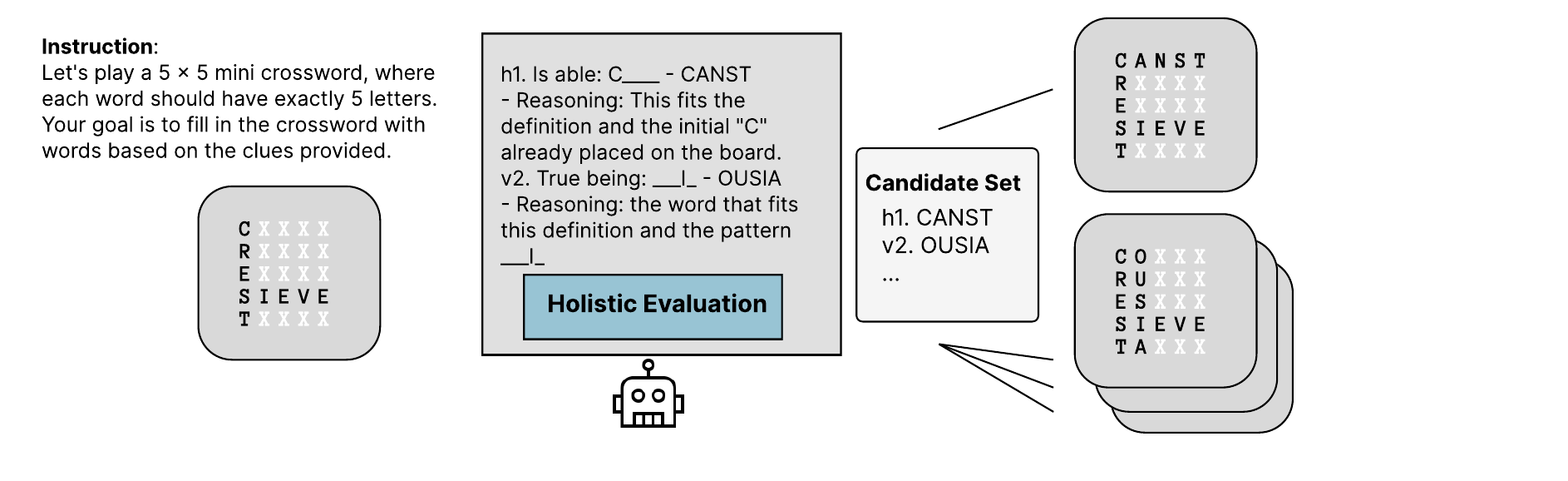}
\caption{\small Illustration of a step in the deliberation process in the Mini-Crosswords task, where the current crossword board is assessed using the thought evaluator and a candidate set of words is proposed for task \ref{task2}. One step is equal to one $d_{rollout}$}
\label{fig:crosswords}
\end{figure*}

\paragraph{Results}

As illustrated in Table \ref{table:outline}, all methods unsurprisingly improve the level of interestingness relative to the initial outline (sampled from default outlines with no prompting for either interesting or boring). However, overall, \methodname{} outperforms ToT even with DFS, while \methodname{} with MCTS demonstrates the highest average interestingness percentage across both GPT-3.5 and GPT-4 with 89.9 and 65.0 respectively.\footnote{One possible explanation for why GPT-4, serving as the base LM, exhibits lower overall interestingness could be attributed to the fact that the outline content evaluator was trained on outlines generated using GPT-3.5.} As Figure \ref{fig:humaneval} shown, human annotators also gave a higher preference towards \methodname{} with MCTS outputs comparing with other baselines, agreeing with the prior evaluation results. Moreover, our strong performance comes at only a modest increase in computational cost compared to baselines.\footnote{We compute the average token cost of \methodname{} for this task along with other tasks in Appendix \ref{appendix:cost-effectiveness}. \methodname{} with DFS has a cost comparable to ToT, while the higher-performing \methodname{} with MCTS requires 1.2x more computation than ToT due to its additional decision simulation process.
}

\paragraph{Continuous Improvement}
Figure \ref{fig:interestingness_cont} illustrates the progression of story outline interestingness at various steps, employing GPT-3.5 as the base LM. Among the tested methods, only \methodname{} with MCTS has exhibited a consistent pattern of improvement over time. In contrast, both ToT and Self-refine exhibit a lack of continuous improvement. We suppose that Self-refine's limited search space and ToT's absence of a revision process contribute to this phenomenon.

\begin{table*}[b]
\centering
\begin{tabular}{@{}lllllll@{}}
\hline
            & \multicolumn{3}{c}{\textbf{GPT3.5}} & \multicolumn{3}{c}{\textbf{GPT4}} \\ \cmidrule(lr){2-4} \cmidrule(lr){5-7}
            
\textbf{Methods}     & \% word     & \% letter     & \% game     & \% word     & \% letter    & \% game    \\ \hline
CoT    &  10.5     &   34.6      &    0.0      &   15.6   &   40.6    &   5.0   \\
Self-refine & 13.5         &  27.4          &    \textbf{5.0}      &   46.5     &   \textbf{74.8}   &   5.0 \\
ToT      & \textbf{19.5}             & 36.6            & 0.0            & 39.5           & 64.8 & 5.0 \\
\methodname{} (DFS)         & 14.0        & 33.2         &    0.0      &   46.5   &    68.2   &  20.0      \\
\methodname{} (MCTS) & 19.0         &  \textbf{41.6}          &     0.0     &      \textbf{54.0}   &  74.0     &  \textbf{25.0}       \\ \hline
\end{tabular}
\caption{\small Mini-crossword results of 20 puzzles for \methodname{} and baselines (success \% of letters, words, and games). \methodname{} with MCTS is either best or closely comparable to best across the board.}
\label{table:crosswordresult}
\end{table*}

\subsection{Mini crosswords}
\label{task2}
We also explore our method on 5x5 mini crosswords following the setup of \citet{yao_tree_2023}. For every puzzle, there are five horizontal (h1 to h5) and five vertical (v1 to v5) words to be filled. The task is to solve a five-by-five crossword puzzle in several steps (either filling or editing a word counts as one step). For evaluation, we check the proportion of letters, words, and games correctly filled by each reasoning method. 

\paragraph{Method Setup}
Each thought node represents a (possibly partial) solution to the crossword puzzle. To evaluate each thought node, the LM is prompted to evaluate each clue against the filled-in letters and suggest whether it is reasonable. For example, if the first row is filled with "AMIGO" and nothing else is filled, then the first column will be shown as "A\_\_\_\_". Thus, in the prompt, there will be one line "v1. A Mennonite sect, named for Jacob Ammann: A\_\_\_\_" that asks the LM to determine whether there are potential answers. The node evaluation's prompt setup is similar to \citep{yao_tree_2023}'s except that we use the evaluation feedback to generate new candidates instead of pruning branches. Based on the evaluation feedback, every candidate for a node will be generated to either suggest a new word to fill a blank space or propose a modification to a word already filled in. For each node, \methodname{} and ToT generate a maximum of 3 candidates. In contrast to the setup in \citet{yao_tree_2023}, where maximum search steps is set to 100, we impose a constraint on all methods to utilize only 20 search steps. This constraint aims to prevent attempts to artificially boost performance by exhaustively trying numerous word possibilities. With this restriction, each row or column of the crossword puzzle allows, on average, only two word attempts to be made within the allocated search budget. Figure \ref{fig:crosswords} illustrates how \methodname{} approaches to solve a crossword puzzle.

\paragraph{Results}
As shown in Table \ref{table:crosswordresult}, \methodname{} with MCTS attains the highest letter success rate using GPT-3.5 and the highest word and game success rate using GPT-4; it is also always at least comparable to the best in all cases. With limited search steps, it is surprising that ToT using GPT-4 performs worse than even Self-refine; it turns out that a self-revision mechanism is important in this task. \methodname{} with MCTS achieves comparable performance to that reported by ToT \citep{yao_tree_2023} using 100 search steps, despite employing just 20 search steps in our experiment.

\subsection{Constrained Generation}
\label{task3}
CommonGen is a benchmark dataset and a constrained text generation task designed to evaluate LMs’ abilities in generative commonsense reasoning \citep{lin-etal-2020-commongen}. An example instruction for the task is shown in Appendix \ref{commengen-prompt}. However, currently, the coverage test of CommonGen can be completed with 90\% or higher accuracy by many LLMs with one-shot prompting. Therefore, we instead test on CommonGen-Hard 
as introduced by \citep{madaan_self-refine_2023}. Rather than just four concepts, CommonGen-Hard requires models to generate a sentence with 20-30 concepts. 

\paragraph{Method Setup}
In this task, we first provide the set of concepts required and the task description for the LM to generate an initial thought node. During the thought evaluation, the LM will be prompted to give feedback about the quality of the concepts used and whether there are any missing concepts. A child node will be generated using the feedback along with the current solution. We set a maximum depth of 3 for this task. For each node, both \methodname{} and ToT will generate a maximum of 3 child candidates. 

\begin{table}[!htbp]
\centering
\begin{tabular}{@{}lll@{}}
\hline
\textbf{Methods}     & \textbf{GPT3.5}  & \textbf{GPT4} \\ 
\hline
CoT    &  44.1  &  96.1  \\
Self-refine &  70.0 &  98.5 \\
ToT & 54.8 & 98.8 \\
\methodname{} (DFS)     &   \textbf{79.6}   &  \textbf{99.1}  \\
\methodname{} (MCTS) &  77.9   &  99.0    \\ 
\hline
\end{tabular}
\caption{\small Constrained Generation Results (\% Coverage of Concepts). \methodname{} outperforms all baselines on both base LMs.}
\label{table:commongen-result}
\end{table}

\paragraph{Results}
Table \ref{table:commongen-result} shows that \methodname{} outperforms all other baselines when using either GPT-3.5 or GPT-4 as the base LM. While \methodname{} with DFS achieves the highest coverage of 79.6\% (GPT-3.5) and 99.1\% (GPT-4), \methodname{} with MCTS also demonstrates comparable concept coverage of 77.9\% using GPT-3.5 and 99.0\% using GPT-4. While MCTS exhibits notable exploration capabilities, it fails to surpass DFS due to the task's nature, where effective solutions are abundant as long as generated sentences correctly integrate assigned concepts. DFS, employing a greedy approach prioritizing nodes with the highest concept coverage, outperforms MCTS in this context. However, solely relying on concept coverage does not ensure appropriate concept utilization. Hence, we conduct an additional evaluation using GPT-4 to determine the preferred output based on concept coverage and appropriateness. Figure \ref{fig:commongen}, comparing \methodname{} with MCTS against \methodname{} with DFS and a third baseline (intuitively, representing the case where neither \methodname{} version's output is good), indicates that \methodname{} with MCTS is significantly favored.

%% file: latex/sections/50-discussion.tex
\section{Conclusion}

We introduce \methodname{}, a framework designed to empower LLMs to handle complex tasks requiring continuous refinement and reasoning capabilities, all without necessitating any modifications or updates to the underlying model architecture.

By harnessing Monte Carlo Tree Search (MCTS), \methodname{} enables LLMs to effectively explore vast search spaces while managing computational resource costs efficiently. Moreover, \methodname{} facilitates a seamless self-revision process, allowing LLMs to iteratively refine and improve their outputs without the need for extensive prompt engineering. Through our experiments, we illustrate \methodname{}'s potential across diverse tasks, highlighting its versatility and broad applicability. The results underscore \methodname{}'s capacity to enhance LLM performance in challenges requiring continuous thought iteration, such as open-ended generation, multi-step reasoning, and creative ideation.

%% file: latex/sections/60-Reproducibility.tex
\section*{Limitations}
While \methodname{} presents a promising approach for reasoning during inference, its reliance on multiple calls to the base language model incurs a higher computational cost than most sampling methods. Consequently, in scenarios where base language models already demonstrate satisfactory performance, the adoption of \methodname{} may not be advisable. However, \methodname{} proves beneficial for tasks requiring intricate reasoning, potential for continual improvement, or when the base language model's performance is suboptimal. Furthermore, the incorporation of MCTS enables \methodname{} to navigate complex search spaces, striking a balance between exploration and exploitation, and handling scalability concerns, thereby offering computational advantages over alternative search algorithms.

\section*{Ethics Statement}
We affirm that all datasets utilized in our experiments have been appropriately sourced and cited, adhering to principles of academic integrity and proper attribution.

Our experiments primarily leverage GPT-3.5 and GPT-4 as the base LLMs. These models possess remarkable capabilities in generating human-like text based on prompts. However, we acknowledge the ethical concerns surrounding their potential misuse for spreading misinformation, generating harmful content, or impersonating individuals. We recognize the imperative for ethical considerations to include robust mechanisms aimed at preventing misuse and fostering responsible use of these models.

The purpose of \methodname{} is to enhance the reasoning and complex problem-solving capabilities of Language Models (LMs). However, it is essential to acknowledge that \methodname{} does not inherently include mechanisms to prevent LMs from generating harmful content. Therefore, we strongly advise anyone utilizing our model to exercise caution and be mindful of the potential for misuse. Users must take proactive measures to mitigate the risk of harmful content generation by implementing effective safeguards and appropriate controls.

\section*{Reproducibility}
In our experiments, we aim for transparency and reproducibility by utilizing publicly accessible datasets. Furthermore, for the content evaluator utilized in the story outline improvement task, we employed Flan-T5, an open-source model. To facilitate reproducibility, our codebase will also be made available for reference and validation upon publication. However, as we access GPT-3.5 and GPT-4 through the OpenAI API, we acknowledge that reproducibility may be affected subject to OpenAI changing their API.

%% file: latex/sections/99-Appendix.tex
\appendix
\input{latex/sections/99-Appendix/A1}
\input{latex/sections/99-Appendix/A2}
\input{latex/sections/99-Appendix/A3}

\input{latex/sections/99-Appendix/A4-outputs-1}

\input{latex/sections/99-Appendix/A4-outputs-3}
\input{latex/sections/99-Appendix/A5-more-evals}
\input{latex/sections/99-Appendix/Additional-experiments}
\input{latex/sections/99-Appendix/weakened_module}

%% file: latex/sections/99-Appendix/A1.tex
\section{Prompts}
\label{sec:appendix-prompt}

Generally, \methodname{} requires only three prompts: TASK\_DESCRIPTION, NEW\_CANDIDATE, and EVALUATE\_CURRENT.
\begin{enumerate}
    \item{TASK\_DESCRIPTION} is the general instruction for the specific task. It will be placed in front of rest of the prompts.
    \item{NEW\_CANDIDATE} is the prompt to generate new candidates based on the evaluation feedback and the current solution. 
    \item{EVALUATE\_CURRENT} instructs the language model to evaluate the current solution. The prompt can be tailored to ask for itemized evaluations, holistic evaluations, or both.
\end{enumerate}

\subsection{Task 1 Story Outline Improvement}

\begin{python}
TASK_DESCRIPTION = """\
# Task Description
You are a popular novel writer. You are now making an interesting outline for the story. You know how to engage with the readers by not limited to introducing interesting characters and unexpected twist.
You also know how to make the story outline coherent and consistent.
"""

NEW_CANDIDATE = TASK_DESCRIPTION + """\
# Original Outline
{outline}

# Feedback
{feedback}

Based on the feedback and the task description, can you make a better story outline by replacing the items suggested by the feedback?

Write the outline in this format just like the original outline from [1] to [{num}]:
[1] ...
[2] ...
...

# Your response:
"""

EVALUATE_CURRENT = TASK_DESCRIPTION + """\
# Original Outline
{outline}

Do you think that this outline is good enough?
Write a score from 1 to 100 where 100 means the outline is perfect based on the task description, and provide an explanation on strengths and weaknesses. Please be specific.
# Write in this format:
[score: 1-100] [reason] xxx (50 words max)

# Example:
[score: 50] [reason] the current outline is too predictable

# Your response:
"""

EVALUATE_CURRENT_ITEMIZED = TASK_DESCRIPTION + """\
Here is a story outline.
{outline}
Which continuous {num_consecutive_lines} outlines items do you think are least interesting?
The interesting outline items should engage readers to read the story. Otherwise, it's boring and should be revised. The interesting level would be from 1 to 5, where 1 is the least interesting and 5 is the most interesting.

Write in this format:
Thought Process:
...
[reason: too repetitive/cliche plot/unsurprising/etc] [start_index]-[end_index] [interesting level: 1-10]

Example:
Thought Process:
Outline items 9 and 10 talks about the same thing over outline items 7 and 8. It's too repetitive.
[reason: too repetitive] [9]-[10] [interesting level: 5]

Can you provide {num_candidates} proposals? 

# Your response:
"""

\end{python}

\paragraph{Fine-tuned story outline content evaluator}
The content evaluator for the story outline is fine-tuned on a Flan-T5 model with a learning rate of 3e-4 and a weight decay of 0.1, trained over 5 epochs.

\begin{python}
PROMPT = """\
Given the story outlines above, do you think that the new story point below is interesting?
"""
\end{python}

\subsection{Task 2 Mini-Crossword Solving}
\begin{python}
TASK_DESCRIPTION = """\
Task Description:
Let's play a 5 x 5 mini crossword, where each word should have exactly 5 letters. Your goal is to fill in the crossword with words based on the hints provided.
"""

NEW_CANDIDATE = TASK_DESCRIPTION + """\
#Current board:
{obs}

#Strategy:
{feedback}

Given the current status of the board and the strategy, list all possible answers for unfilled or changed words, and your confidence levels (certain/high/medium/low), using the format like this:
Use "certain" cautiously and only when you are 100

h1. [hint: _____] xxxxx (medium)
h2. [hint: _____] xxxxx (certain)
...
v1. [hint: _____] xxxxx (high) 
...

Write your response in the format:
h1. [A financial loss; a negative profit; to remove bits from: D_B__] DEBTS (low)
h2. [Fatuous; empty headed: _____] INANE (high)
...
v1. [A dice player; something that cuts into small cubes: _____] DICER (high)
v5. [An Indian tent: _____] TEPEE (medium)

Each line can only have one candidate answer. 
#Your response:
"""

EVALUATE_CURRENT = TASK_DESCRIPTION + """\
# Current board:
{obs}
Evaluate the current board and provide a strategy on how to continue to fill in the blank or correct potential mistakes.
Write your response in the format:
v1. [reasoning and potential answers] 
v2. [reasoning and potential answers] 
...
h1. [reasoning and potential answers]
...
# Example:
v2. [Current answer: tough; since  the filled in h1. is debit; e is conflicted with t, we could consider other options such as ENURE]
v3. [Current answer: ??? CUTUP could be a potential answer]
# Your response:
"""
\end{python}

\subsection{Task 3 Constrained Generation}
\label{commengen-prompt}
\begin{python}
TASK_DESCRIPTION = """\
# Instruction Given several concepts (i.e., nouns or verbs), write a short and simple sentence that contains *all* the required words. The sentence should describe a common scene in daily life, and the concepts should be used in a natural way. 
# # Examples 
# ## Example 1 - Concepts: "dog, frisbee, catch, throw" - Sentence: The dog catches the frisbee when the boy throws it into the air. 
# ## Example 2 - Concepts: "apple, place, tree, pick" - Sentence: A girl picks some apples from a tree and places them into her basket. 
"""
INSTRUCTION = """\
Your Task - Concepts: {concepts}
"""
NEW_CANDIDATE = TASK_DESCRIPTION + """\
Instruction:
{instruct}

Here is a proposed sentence.
{solution}

Here is the feedback of outline item.
{feedback}

Based on the feedback, can you make a revised solution?
# Sentence:
"""
EVALUATE_CURRENT = TASK_DESCRIPTION + """\
Instruction:
{instruct}

Here is a proposed sentence.
{solution}

Do you think that the proposed sentence is good enough? Write "no need to improve" if you think 1) the sentence covers all the concepts listed in the instruction; and 2) the sentence describes a common scene in daily life.

Otherwise, write "still need to improve" and provide a reason.

# Write in this format:
[No need to improve/still need to improve] [reason] xxx (50 words max)

# Example 1:
[still need to improve] the sentence misses the concept "dog", "ladder", and "drum".
# Example 2:
[still need to improve] the cat does not fly.

# Your response:
"""
\end{python}

%% file: latex/sections/99-Appendix/A2.tex
\newpage
\section{Computation Efficiency}
\label{appendix:cost-effectiveness}
Table \ref{table:cost}, Table \ref{table:cost2}, and Table \ref{table:cost3} show the estimated number of input/output tokens usage and the cost of completing one case.
\methodname{} with DFS has a comparable cost to ToT while \methodname{} with MCTS requires a greater computation since it has an additional decision simulation process.

\begin{table*}[!htbp]
\centering
\begin{tabular}{lll}
\hline
               & Input/Output Tokens & Cost per case \\ \hline
ToT            & 10.1k/4.9k          & \$0.248       \\
\methodname{} with DFS  & 11.3k/4.6k          & \$0.251      \\
\methodname{} with MCTS & 25.0k/9.9k          & \$0.547      \\ \hline
\end{tabular}
\caption{Token use and estimated cost for Story Outline Improvement (Base LLM: gpt-4-0125-preview)}
\label{table:cost}
\end{table*}

\begin{table*}[!htbp]
\centering
\begin{tabular}{lll}
\hline
             & Input/Output Tokens & Cost per case    \\ \hline
ToT             & 64.5k/8.9k          & \$0.912        \\
\methodname{} with DFS   & 41.6k/7.1k          & \$0.629       \\
\methodname{} with MCTS   & 100.2k/16.3k        & \$1.491       \\ \hline
\end{tabular}
\caption{Token use and estimated cost for Mini-Crossword (Base LLM: gpt-4-0125-preview)}
\label{table:cost2}
\end{table*}

\begin{table*}[!htbp]
\centering
\begin{tabular}{lll}
\hline
              & Input/Output Tokens     & Cost per case    \\ \hline
ToT                   & 7.1k/1.1k               & \$0.104          \\
\methodname{} with DFS         & 7.0k/0.7k               & \$0.091          \\
\methodname{} with MCTS        & 15.7k/2.0k              & \$0.217          \\ \hline
\end{tabular}
\caption{Token use and estimated cost for Constrained Generation (Base LLM: gpt-4-0125-preview)}
\label{table:cost3}
\end{table*}

%% file: latex/sections/99-Appendix/A3.tex
\newpage
\section{Alternative Search Algorithm}
\begin{algorithm}
\label{mcts_code}
\caption{\methodname{} with MCTS}\label{alg:mcts}
\begin{algorithmic}[1]
\STATE \textbf{Input:} Initial node $x_0$
\STATE \textbf{Output:} Output node $x^{*}$
\STATE Initialize empty search tree $T$
\FOR{$j \leftarrow 1$ to $d_{rollout}$}
    \STATE Select a leaf node $x$ using the tree policy UCB1 Eqn \ref{eq:UCB1}
    \STATE Expand node $x$ by generating a set of children nodes $X_{\text{child}}$
    \STATE node $x \leftarrow \text{uniformly\_sampled} (X_{\text{child}})$
    \FOR{$k \leftarrow 1$ to $d_{simulation}$}
        \STATE node $x \leftarrow \text{generate\_single\_child} (x)$
    \ENDFOR
    \STATE Evaluate reward $v(x)$
    \STATE Propagate the reward $v$ and number of explorations $n$ back to $x_0$
\ENDFOR
\STATE Choose the best node $x^{*}$ with the highest reward $v$
\RETURN $x^{*}$
\end{algorithmic}
\end{algorithm}

\begin{algorithm}
\caption{\methodname{} with DFS}\label{alg:dfs_code}
\begin{algorithmic}[1]
\STATE \textbf{Input:} Initial node $x$, Depth $d$
\STATE \textbf{Output:} Goal node $x^{*}$
\STATE $x \leftarrow x_0$
\IF{$d = 0$}
    \RETURN $x$
\ENDIF

\STATE Expand node $x$ by generating a set of children nodes $X_{\text{child}}$
\FOR{$k \leftarrow 1$ to $max\_candidates$}
    \STATE Evaluate reward $v(X_{child}[k])$
\ENDFOR

\STATE Choose the node $x^{*}$ with the highest reward $v$ in $X_{child}$
\STATE $DFS(x^{*}, d - 1)$

\end{algorithmic}
\end{algorithm}

%% file: latex/sections/99-Appendix/A4-outputs-1.tex
\newpage
\section{Output Examples}
\subsection{Story Outline Improvement}
The examples below show the final outputs of each reasoning algorithm in the story outline improvement task. In general, \methodname{} with MCTS displays a more complex story with a greater tendency to introduce twists and characters. The final output of \methodname{} with MCTS may be greatly modified. Self Refine is less prone to change from the initial outline. ToT and \methodname{} with DFS are in the middle and would like to add interesting details to the initial outlines.

\subsubsection{Example 1} 

Initial Outline:
\begin{enumerate}

\item Jack arrives in Maine and starts attending the same high school as Leo.
\item Jack begins observing Leo's daily routine and interests to find common ground.
\item Jack strategically plans his first encounter with Leo to establish a connection.
\item Jack successfully befriends Leo and starts spending more time together.
\item Jack and Leo spend more time together, deepening their bond.
\item Jack and Leo's attraction becomes undeniable, leading to their first kiss.
\item Jack and Leo navigate the challenges of their secret relationship.
\item Jack and Leo's love for each other grows stronger, despite the risks involved.
\item The Center orders Jack to cut ties with Leo and leave Maine immediately.
\item Jack struggles with the decision to leave Leo and return to the Center.
\item Leo's father becomes a target of the Center's operation, putting Leo in danger.
\item Jack risks his own safety to protect Leo and his father, defying the Center's orders.

\end{enumerate}

Self Refine:
\begin{enumerate}
\item Jack arrives in Maine and starts attending the same high school as Leo.
\item Jack begins observing Leo's daily routine and interests to find common ground.
\item Jack strategically plans his first encounter with Leo to establish a connection.
\item Jack successfully befriends Leo and starts spending more time together.
\item Jack and Leo spend more time together, deepening their bond.
\item Jack and Leo's attraction becomes undeniable, leading to their first kiss.
\item Jack and Leo navigate the challenges of their secret relationship.
\item Jack and Leo's love for each other grows stronger, despite the risks involved.
\item The Center orders Jack to cut ties with Leo and leave Maine immediately.
\item Jack struggles with the decision to leave Leo and return to the Center.
\item A mysterious organization threatens to expose Jack and Leo's relationship, putting their lives in danger.
\item Jack must make a decision to protect Leo and himself, risking everything to defy both the Center and the new threat.
\end{enumerate}

ToT:
\begin{enumerate}
\item Jack arrives in Maine and starts attending the same high school as Leo.
\item Jack is immediately drawn to Leo's mysterious and aloof demeanor, sparking his curiosity.
\item Jack discovers Leo's passion for art and secretly starts leaving anonymous gifts for him.
\item Leo becomes intrigued by the gifts and begins to search for the mysterious admirer.
\item Jack and Leo finally meet at a school art exhibition, where Leo realizes Jack's connection to the gifts.
\item Their initial awkward encounter slowly blossoms into a genuine friendship as they bond over their shared love for art.
\item Jack and Leo's friendship is tested when a scandal rocks their school, threatening to tear them apart.
\item Despite the challenges, Jack and Leo's bond deepens as they confide in each other and support one another.
\item The Center, a secretive organization Jack is a part of, reveals their true intentions and orders Jack to complete a dangerous mission that puts Leo at risk.
\item Jack is torn between his loyalty to the Center and his love for Leo, struggling to find a way to protect both.
\item Leo's life is put in grave danger when he uncovers a dark secret that implicates the Center in a series of crimes.
\item Jack makes the ultimate sacrifice to save Leo and expose the Center's corrupt actions, putting his own life on the line for the person he loves.

\end{enumerate}

\methodname{} with DFS
\begin{enumerate}
\item Jack arrives in Maine and starts attending the same high school as Leo.
\item Jack is immediately drawn to Leo, but soon realizes that Leo is hiding a dark secret.
\item Instead of confronting Leo right away, Jack decides to secretly follow him and discovers that Leo is actually working undercover to bring down the dangerous organization.
\item Jack is conflicted between his growing feelings for Leo and the danger of being involved in Leo's mission.
\item Jack offers to help Leo in his mission, leading to a deepening of their bond as they face increasing risks together.
\item Their forbidden romance blooms amidst the chaos, complicating their mission and putting them both in jeopardy.
\item The organization catches wind of Jack and Leo's alliance, putting both of their lives at risk.
\item Jack and Leo must make difficult choices to protect each other and navigate the treacherous waters of their relationship.
\item The Center threatens to harm Leo's family if they don't stop their investigation.
\item Jack and Leo are forced to go underground and make a dangerous deal to protect their loved ones.
\item The stakes are raised when Leo's father is kidnapped, pushing Jack and Leo to their limits in a race against time to save him.
\item Jack and Leo face off against the leaders of the organization in a thrilling final showdown, risking everything to bring them down and secure a future together.
\end{enumerate}

\methodname{} with MCTS:
\begin{enumerate}
\item Jack arrives in Maine and starts attending the same high school as Leo.
\item Jack notices a mysterious new student, Alex, who seems to be watching Leo closely.
\item Jack becomes curious about Alex's intentions and starts investigating their background.
\item Jack discovers that Alex is actually a spy from a rival organization called the Center targeting Leo.
\item Jack confronts Alex and warns Leo about the danger, leading to a confrontation between the three.
\item Jack, Leo, and Alex form an unlikely alliance to uncover the truth behind the Center's motives.
\item Jack and Leo's bond is tested as they navigate the complexities of trust and betrayal within their group.
\item The rival organization escalates their actions, putting Jack, Leo, and Alex in grave danger.
\item Leo sacrifices himself to protect Jack and defeat the Center, shocking everyone.
\item The truth behind the Center's motives is revealed, leading to a deeper understanding of the sacrifices made.
\item Jack and Alex must work together to honor Leo's sacrifice and take down the Center.
\item In a shocking revelation, it is uncovered that Leo's twin brother, who was thought to be dead, is actually the mastermind behind the Center, creating a complex dynamic between the characters. 
\end{enumerate}

\subsubsection{Example 2}
Initial outline:
\begin{enumerate}
\item The Health Department wants to close the Boys Club due to infestation of rats and cockroaches.
\item Closing the Boys Club means the kids will be left without a place to go.
\item The kids are worried about being out on the street again.
\item Buddy suggests organizing a bake sale to raise funds.
\item Johnny proposes hosting a talent show as another fundraising idea.
\item Rich suggests reaching out to local businesses for donations or sponsorships.
\item Cool suggests organizing a community car wash to raise money.
\item Buddy Williams suggests asking the local pet store owner if he knows anything about the rich cat.
\item The kids visit the pet store and ask the owner about the rich cat.
\item The pet store owner tells the kids about a rumor he heard of a rich cat living in a mansion nearby.
\item The kids decide to investigate the mansion and see if they can find the rich cat.
\item The townspeople share their findings and clues with the kids, who compile the information and continue the search.
\item The kids and townspeople follow leads and investigate potential sightings of the rich cat.
\item The search intensifies as more people join in, creating a sense of community and determination to find the rich cat.
\item After an extensive search, the kids and townspeople finally locate the rich cat, who generously donates the money needed to save the Boys Club.
\end{enumerate}
Self Refine:
\begin{enumerate}
\item The Health Department wants to close the Boys Club due to infestation of rats and cockroaches.
\item Closing the Boys Club means the kids will be left without a place to go.
\item The kids are worried about being out on the street again.
\item Buddy suggests organizing a bake sale to raise funds.
\item Johnny proposes hosting a talent show as another fundraising idea.
\item Rich suggests reaching out to local businesses for donations or sponsorships.
\item Cool suggests organizing a community car wash to raise money.
\item Buddy Williams suggests asking the local pet store owner if he knows anything about the rich cat.
\item The kids visit the pet store and ask the owner about the rich cat.
\item The pet store owner tells the kids about a rumor he heard of a rich cat living in a mansion nearby.
\item The kids decide to investigate the mansion and see if they can find the rich cat.
\item While searching the mansion, the kids stumble upon a hidden room filled with mysterious artifacts and clues.
\item The artifacts lead them to a map that hints at a hidden treasure in the town.
\item The kids and townspeople embark on a thrilling treasure hunt, following the map's clues and facing challenges along the way.
\item After overcoming obstacles and working together, the kids and townspeople unearth the hidden treasure, which turns out to be the rich cat's fortune that saves the Boys Club.
\end{enumerate}

ToT:
\begin{enumerate}
\item The Health Department wants to close the Boys Club due to infestation of rats and cockroaches.
\item Closing the Boys Club means the kids will be left without a place to go.
\item The kids are worried about being out on the street again.
\item Buddy suggests organizing a bake sale to raise funds.
\item Johnny proposes hosting a talent show as another fundraising idea.
\item Rich suggests reaching out to local businesses for donations or sponsorships.
\item Cool suggests organizing a community car wash to raise money.
\item Buddy Williams suggests asking the local pet store owner if he knows anything about the rich cat.
\item The kids visit the pet store and ask the owner about the rich cat.
\item The pet store owner tells the kids about a rumor he heard of a rich cat living in a mansion nearby.
\item The kids decide to investigate the mansion and see if they can find the rich cat.
\item While investigating the mansion, the kids stumble upon a hidden room with a mysterious map that leads to a hidden treasure.
\item The townspeople share their findings and clues with the kids, who now have a new mission to find the hidden treasure.
\item The search for the hidden treasure leads the kids and townspeople to uncover secrets about the town's history and a long-lost treasure that could save the Boys Club.
\item After a thrilling adventure and overcoming obstacles, the kids and townspeople finally find the hidden treasure, which not only saves the Boys Club but also brings the community closer together.
\end{enumerate}

\methodname{} with DFS
\begin{enumerate}
\item The Health Department discovers that the Boys Club is in danger of losing its funding due to budget cuts.
\item The kids are devastated as they learn that the Boys Club may have to shut down if they can't come up with a solution.
\item Buddy suggests organizing a scavenger hunt around town to raise funds and awareness for the Boys Club.
\item Johnny proposes creating a social media campaign to rally support and donations for the Boys Club.
\item Rich suggests hosting a charity concert with local musicians and performers to raise money.
\item Cool suggests organizing a charity auction with items donated by local businesses and community members.
\item The kids brainstorm and come up with a plan to combine all the fundraising ideas into one big event to save the Boys Club.
\item Buddy Williams suggests asking the local historian if he knows of any hidden treasures or artifacts that could help raise funds.
\item The kids visit the local historian and learn about a legendary treasure rumored to be hidden somewhere in town.
\item The historian gives the kids a map and clues to help them on their quest to find the treasure.
\item The kids embark on an exciting adventure to follow the map and solve the clues to locate the hidden treasure.
\item Along the way, they face challenges and obstacles that test their friendship and determination.
\item The townspeople join in the treasure hunt, adding to the excitement and suspense of the search.
\item After overcoming obstacles and following the clues, the kids and townspeople finally discover the hidden treasure, which turns out to be the key to saving the Boys Club.
\item The treasure not only provides the funds needed to keep the Boys Club open but also brings the community together in a heartwarming and unforgettable way.
\end{enumerate}

\methodname{} with MCTS

\begin{enumerate}
\item The Health Department wants to close the Boys Club due to infestation of rats and cockroaches.
\item Closing the Boys Club means the kids will be left without a place to go.
\item The kids are worried about being out on the street again.
\item Buddy suggests organizing a scavenger hunt around town to raise funds.
\item Johnny proposes creating a community mural project as another fundraising idea.
\item Luna, a mysterious new girl in town, suggests hosting a talent show to raise money, revealing her own hidden talents in the process.
\item Rich suggests organizing a neighborhood cleanup event to raise money.
\item Cool suggests hosting a movie night in the park to raise funds.
\item Buddy Williams suggests asking the local historian if he knows anything about a hidden treasure in the town.
\item The kids visit the local historian and ask about the hidden treasure.
\item The historian tells the kids about a legend of a lost pirate treasure buried somewhere in the town.
\item Luna shares a personal connection to the treasure hunt, sparking curiosity and excitement among the group.
\item The kids face challenges and puzzles during their treasure hunt that test their problem-solving skills and teamwork.
\item In a surprising twist, the treasure turns out to be a time capsule filled with memories and mementos from the town's past.
\item The kids realize that the true treasure was the journey itself and the bonds of friendship they formed along the way, teaching them the value of community and unity. Luna's involvement adds depth to the story and enhances the theme of unity and friendship.
\end{enumerate}

%% file: latex/sections/99-Appendix/A4-outputs-3.tex
\newpage
\subsection{Constrained Generation}
The examples below show the final outputs of different reasoning strategies in Constrained Generation.

\subsubsection{Example 1}
\textbf{Concepts}: drum, clip, window, cover, graffiti, wave, vomit, hurdle, dye, sidewalk, animal, slice, swing, pottery, pineapple, fix, decorate, tire, tag, bicycle, watermelon, sign, mask, plant

\paragraph{Refine}

Children decorated pottery with vibrant dye near the graffiti-covered wall, while a masked animal rode a bicycle past a sign, waving as they passed by a tire swing hanging from a tree and a clip holding a sign on the sidewalk, avoiding a hurdle, slicing a pineapple and a watermelon to plant in the garden.

\paragraph{ToT}

A child on a swing by the sidewalk waved at a bicyclist with a tire fixed and decorated with graffiti.

\paragraph{\methodname{} with DFS}

The children decorated the pottery with vibrant dye, while the adults fixed the tire on the bicycle near the sidewalk, as a group of friends played the drum and waved at passersby from the swing, with a clip holding a sign on the window, a slice of watermelon in hand, and a pineapple plant in the background.

\paragraph{\methodname{} with MCTS}

A person walks down the sidewalk past a graffiti-covered window, a pottery clip lies on the ground, a bicycle rider swerves to avoid a tire, and a sign is being fixed and decorated with dye next to a pineapple stand, as a drum beats in the background, a wave crashes on the shore, an animal scurries across the path, a slice of watermelon is enjoyed under the shade of a swing, a mask-wearing individual tags a plant with a new growth, and a hurdle is cleared by a runner as they vomit from exhaustion.

\subsubsection{Example 2}
\textbf{Concepts}: plant, word, remove, toddler, hamburger, potato, set, line, place, control, eat, banner, singe, hurdle, bed, bar, alley, stove, top, air, cowboy, sharpen, pair, beat, arena, wave, singer, jacket

\paragraph{Refine}

The toddler removes the banner from the line and places it on the top of the bar to singe while the cowboy sharpens a pair of knives to beat the hurdle in the arena.

\paragraph{ToT}

The toddler eats a hamburger and potato at the place where the plant is growing, while the singer on stage waves under the banner to the cheering crowd, and the cowboy sharpens his pair of beat-up boots backstage.

\paragraph{\methodname{} with DFS}

The toddler eagerly eats a hamburger while the cowboy sharpens his pair of scissors at the plant nursery.

\paragraph{\methodname{} with MCTS}

The toddler removes a plant from the line, places it in the bed, eats a hamburger with a side of potato, then sets the table for a cowboy sharpening his pair of knives at the top of the bar, while a singer controls the banner in the arena, singing as the wave of music beats through the alley where a jacketed cowboy hurdles over a singe stove.

%% file: latex/sections/99-Appendix/A5-more-evals.tex
\newpage
\section{G-Eval on Story Outline Generation}
We’ve run an additional evaluation using the G-Eval metric \cite{liu2023gevalnlgevaluationusing}. We provide a definition prompt of interestingness, and the result indicates that ThoughtSculpt(MCTS) outperforms other baselines when evaluated by either GPT-4 and GPT-3.5, consistent with other metrics.

\begin{table}[h!]
\centering
\begin{tabular}{@{}lll@{}}
\hline
 & \textbf{GPT-3.5} & \textbf{GPT-4} \\ 
\hline
Self-refine & 4.33 & 4.45 \\ 
ToT & 4.37 & 4.66 \\ 
ThoughtSculpt (DFS) & 4.47 & 4.71 \\ 
ThoughtSculpt (MCTS) & 4.60 & 4.73 \\ 
\hline
\end{tabular}
\caption{G-Eval Result (1-5 scale)}
\label{table:geval}
\end{table}

\section{More search steps on Mini-Crosswords}
We've shown that \methodname{} (MCTS) could achieve a solid performance in only 20 search steps, but we have run an extended number of search steps to match the experiment setup provided by \citet{yao_tree_2023}.

\begin{table*}[h]
\centering
\begin{tabular}{@{}llll@{}}
\toprule
             & \multicolumn{3}{c}{\textbf{GPT4}} \\ \hline
            
Methods    & \% word     & \% letter    & \% game    \\ 
\midrule
ToT (20 search steps)      & 39.5           & 64.8 & 5.0 \\
ToT (100 search steps) \cite{yao_tree_2023}  &     60   & 78 & 20 \\
\methodname{} (MCTS) (20 search steps)   &      54.0  &  74.0     &  25.0      \\ 
\methodname{} (MCTS)  (100 search steps)       &   \textbf{66.0}   &    \textbf{83.0}   &  \textbf{35.0}      \\ \hline
\end{tabular}
\caption{\small Mini-crossword results of 20 puzzles for \methodname{} and baselines (success \% of letters, words, and games). Comparison between ToT and \methodname{} (MCTS) with 20 and 100 search steps}
\label{table:crosswordresult}
\end{table*}

\section{Constrained Generation LLM Evaluation}
\begin{figure*}[htbp]
    \centering
    \begin{subfigure}[b]{0.4\textwidth}

        \includegraphics[width=\textwidth]{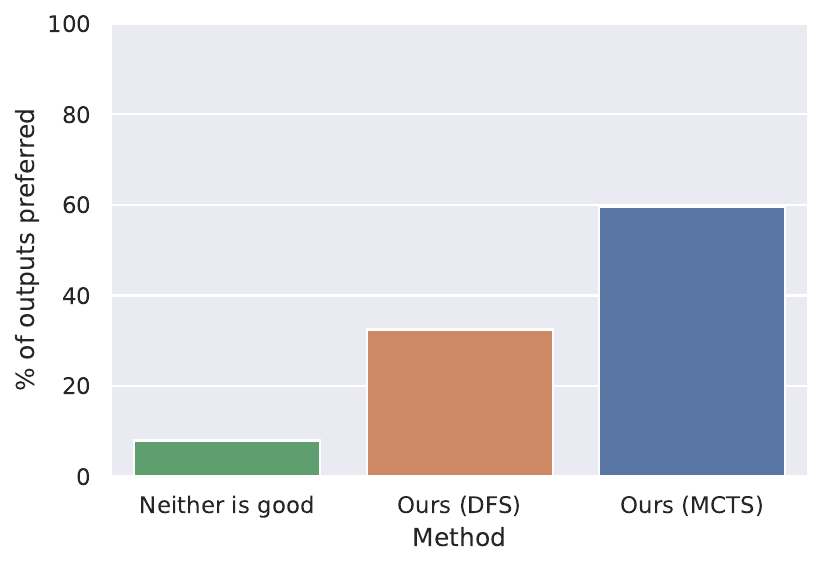}
        \caption{\small Base LLM GPT-3.5}
        \label{fig:sub1}
    \end{subfigure}
    \begin{subfigure}[b]{0.4\textwidth}
        \includegraphics[width=\textwidth]{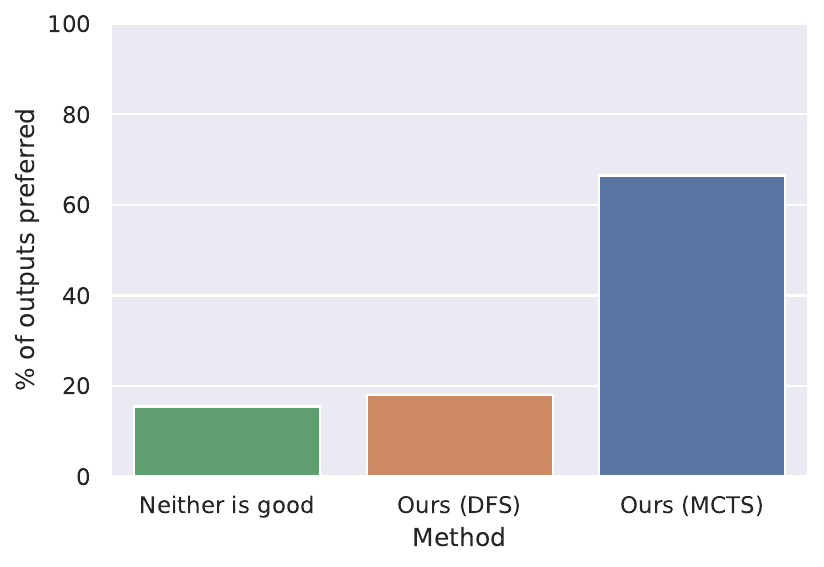}
        \caption{\small Base LLM GPT-4}
        \label{fig:sub2}
    \end{subfigure}
    \caption{\small GPT-4's comprehensive preference based on concept coverage and appropriateness over the final outputs for Constrained Generation. \methodname{} with MCTS is preferred by a wide margin.}
    \label{fig:commongen}
\end{figure*}

To evaluate the preferred output based on concept coverage and appropriateness, we conduct an additional assessment using GPT-4. We prompt GPT-4 to select the output that is most preferred, considering both its coverage of relevant concepts and the appropriateness of how this coverage is utilized. Figure \ref{fig:commongen}, which compares \methodname{} with MCTS against \methodname{} with DFS and a third baseline (which intuitively represents the case where neither version of \methodname{} produces a satisfactory output), shows that \methodname{} with MCTS is significantly favored.

%% file: latex/sections/99-Appendix/Additional-experiments.tex
\newpage
\section{Game of 24}
We additionally experiment with our method on the game of 24 using the same test set provided by \citep{yao_tree_2023}. Compared to \citep{yao_tree_2023}'s prompts with detailed few-shot examples for this task, \methodname{} uses a much more general prompt. As illustrated in Table \ref{table:gameof24}, \methodname{} still outperforms ToT in this comparison despite this setup that might be expected to favor ToT.

\begin{table}[h!]
\centering
\begin{tabular}{@{}lll@{}}
\toprule
 & \textbf{Success \%} \\
\midrule
CoT-SC (k=100) \cite{yao_tree_2023} & 9 \\ 
IO - Refine (k=10) \cite{yao_tree_2023}	 & 27  \\ 
ToT (b = 1) \cite{yao_tree_2023}	& 45  \\ 
ToT (b = 5) \cite{yao_tree_2023}	& 74  \\
ThoughtSculpt (MCTS) (b = 5) & 79  \\ 
\hline
\end{tabular}
\caption{Game of 24 Result }
\label{table:gameof24}
\end{table}

\subsection{Game of 24 Task prompts}
\begin{python}
TASK_DESCRIPTION = """\
Use the given four numbers and basic arithmetic operations (+ - * /) to obtain 24. 
You can use the numbers only once but you can use them in any order.
"""

SOLUTION_OUTPUT_FORMAT = """\
# Think step by step first. Then, please output the solution in the following format (in a Python code block).
```python
(1 + 2) * (2 * 4)
```
# Your response. 
"""

REVISE_SOLUTIONS = """\
# Instruction
{instruction}

# Current Solution
{solution}

Calculate the result of the current solution.
Do you think the solution is correct? If not, please provide feedback. 

# Output format

```json
{{
    "calculation": "step by step calculation of the current solution",
    "result: int,
    "feedback": "Your feedback here",
    "correct" : true/false
}}
```
"""

\end{python}

%% file: latex/sections/99-Appendix/weakened_module.tex
\newpage
\section{Effectiveness of components}
Our framework has two core components that use LLM: the thought evaluator, providing both numeric and language feedback for revision, and the thought generator, which generates subsequent nodes based on this feedback to improve outputs. Both are essential, as removing either makes the method non-functional. We conducted an additional ablation study by "weakening" these modules. We replaced GPT-4 with Llama3.1-8B-Instruct as the language model for either the thought evaluator or the thought generator in the Constrained Generation task. The results are summarized below:

\begin{table}[h!]
\centering
\begin{tabular}{@{}lll@{}}
\toprule
\textbf{Modules} & \textbf{\% Coverage} \\
\midrule
ThoughtSculpt & 99.0 \\
ThoughtSculpt (weakened generator) & 98.5 \\
ThoughtSculpt (weakened evaluator) & 96.3 \\
ThoughtSculpt (weakened generator + evaluator) & 90.7 \\
\bottomrule
\end{tabular}
\caption{Coverage of Different ThoughtSculpt's Components}
\end{table}

These results highlight the critical role of a strong evaluator in deep reasoning tasks, as it provides essential revision feedback that significantly enhances the generator’s effectiveness. 